\documentclass{article}
\usepackage{PRIMEarxiv}
\usepackage{bm}
\usepackage[utf8]{inputenc} 
\usepackage[T1]{fontenc}    
\usepackage{hyperref}       
\usepackage{url}            
\usepackage{booktabs}       
\usepackage{amsfonts}       
\usepackage{nicefrac}       
\usepackage{microtype}      
\usepackage{lipsum}
\usepackage{graphicx}
\graphicspath{{fig/}}     
\usepackage{tikz}
\usepackage{calc}
\usepackage{algorithm, algorithmic}
\usepackage{amsmath}
\usepackage{graphicx}  
\usepackage{float}  
\usepackage{subfigure}  

\title{Mesh-SORT: Simple and effective location-wise tracker with lost management strategies
}

\author{
  ZongTan Li \\
  Maynooth International Engineering College \\
  FuZhou University \\
  FuZhou, China\\
  \texttt{zongtanli2023@163.com} \\
}

\begin{document}
\maketitle

\begin{abstract}
    Multi-Object Tracking (MOT) has gained extensive attention in recent years due to its potential applications in traffic and pedestrian detection.  We note that tracking by detection may suffer from errors generated by noise detectors, such as an imprecise bounding box before the occlusions, and observed that in most tracking scenarios, objects tend to move and lost within specific locations.   To counter this, we present a novel tracker to deal with the bad detector and occlusions. Firstly, we proposed a location-wise sub-region recognition method that equally divided the frame, which we called mesh. Then we proposed corresponding location-wise loss management strategies and different matching strategies. The resulting Mesh-SORT, ablation studies demonstrate its effectiveness and made 3\% fragmentation 7.2\% ID switches drop and 0.4\% MOTA improvement compared to the baseline on MOT17 datasets.  Finally, we analyze its limitation on the specific scene and discussed what future works can be extended.

\end{abstract}


\section{Introduction}
Multi-object tracking (\textbf{MOT}) is a challenging task in computer vision. Its main focus is on tracking determined objects by a given video or frame sequence. It can benefit intelligent traffic analysis and human-computer interaction. There are two dominant categories of Multiple Object Tracking (MOT) methods: online and offline. Offline methods generate trajectories using information from both past and future frames. In contrast, online methods only use information from the current and previous frames. While offline methods can better handle certain ambiguous tracking issues, they are not capable of performing real-time vision tasks. Since real-time performance is essential for pedestrian tracking systems, this study focuses on online MOT. 

In online tracking, recent studies show that two scheme are reached considerable progress: end-to-end and tracking-by-detection(TBD)\cite{zhang2022motrv2,cao2022ocsort, SMILEtrack}.  end-to-end paradigm always use a single module to solve the problem, which needs complex structure design and excellence training techniques.  While tracking by detection scheme generally divided this problem into two models, which are performing \textit{detection} for each frame and then doing the \textit{association} for combining objects. Where the association is including two main parts: 1) Motion and state forecast. Because only bounding boxes information is provided to the association part, dynamics and motion may lose during this process, the Kalman filter provides a solution for linear motion prediction. If non-linear motion happens, the inaccurate prediction will make association unable under the complex scene.  2) Matching the predicted tracklet and new frame detection using the designed distance metrics which represent spatiotemporal consistency (e.g IoU, ReID, momentum\cite{cao2022ocsort}, Cascaded buffer IoU \cite{yang2023hard}).  The main issues of TBD scheme now being addressed are:

1) Frame scene comprehension and pattern understanding. Extra information on the frame can be utilized and applied to the association part, for example, Re-ID module could be used to extract the object-wise feature in the scene.
2) Dealing with unsuccessful detector proposals. Because detectors can only provide information, how recognizing bad detection is also an essential part of tracking.
3) Finding the spatiotemporal similarity and consistency matching policy of tracking objects for occlusion and non-linear motion. The improvement mainly focus on the spatiotemporal prior to the bounding box. Find the maximum probability of matching. 


 Objects in real scenes often have varying distances from the camera, resulting in bounding boxes generated by the detector that vary in size depending on the object. Additionally, if two objects have similar velocities, their movement in the camera view may not have the same magnitude. Therefore, spatio information can also be encoded into module for tracking, the CorrTraker\cite{corrtracker} provides a module to adaptively learn the object's feature correlation. And AOH\cite{AOH} proposed a density estimator module and potential object mining module to deal with crowded scene occlusion using global information. Through extensive observations, we have found that the differences due to camera perspective are location-dependent in most tracking scenarios, which gives us the inspiration to solve this problem.

At the same time, the detector proposal with the noise made different levels of impact on the tracker. Currently, most SORT-like methods can not build a feedback mechanism between detection and association. We found that the detector may play badly in specific situations, especially in occlusion, this may cause bad performance in the Kalman filter.
To address the issues above, we propose a location-wise tracker that has an adaptive law to track the object. Our work can be summarized below:
Firstly, we proposed a general location-wise framework in tracking by detection scheme in multi-object tracking, divided the sub-region of the frame by mesh, and identify the crowded region with the  number of lost tracklet. 
Second, we adopt the lost tracklet management strategies combined with mesh adaptive strategies.  We proposed a lost maintenance (LM) mechanism to maintain the tracklet to increase tracking consistency and reduce fragmentation.
Third, we made a derivation of detector noise before the occlusions, and we exploit the saved velocity to tackle this issue. A number of ablation experiments has been performed to prove the effectiveness of each contribution of the proposed method.
\section{Related Works}
\label{sec:headings}


\subsection{Tracking-by-detection}
With wide application of profound methods in object detection, the tracking-by-detection can benefit from more accurate detection.  It breaks the tracking problem into two stages process, detection, and association.  Many past state of art methods followed this paradigm and reached high performance. SORT\cite{bewley2016simple} is the most popular motion-based method which inspired many SORT-like work after it  was proposed.
  DeepSORT\cite{wojke2017simple} introduced Re-ID to compensate for the detector incapacity by changing the distance metric with an appearance feature. However, Re-ID may fall short in a crowded scene, and the high computational effort and the large size of the required training set make this module significantly less versatile. Recent methods in this scheme are more focused on refining adjustments on the appearance and geometric consistentency  for matching score and enhancing matching strategies. For example, C-IoU tracker\cite{yang2023hard} exploit the buffer IoU to mitigate the irregular motion and cross frame affinity, StrongSORT\cite{du2022strongsort} proposed the Guassian-smoothed interpolation to enhance the trajectory consistency for the bad detection. Also, there are many works are done on spatial information, Correlation Tracker\cite{corrtracker} proposed the module for extracting spatial correlation between objects, then doing the self-supervised learning to construct the online context map.
For past state-of art method,  BoT-SORT\cite{aharon2022bot} modified the state of Kalman filter and ReID's cosine-distance fusion for more robust association between detections and tracklet.
While OC-SORT\cite{cao2022ocsort} assume that the detector will give more faithful results, the tracking scheme reuses the observation before developing three modules to improve accuracy, ByteTrack\cite{zhang2021bytetrack} states that confidence generated by the detector will decrease when the targets are occluded, so an improvement on low threshold can find the occlusion target back. improving work like BoT-SORT focuses on modeling camera motion on the video \cite{aharon2022bot}.

\subsection{Other Schemes}
\paragraph{Detection Embedding}
The detection embedding-based method estimates the bounding box and embeds the target's feature on another module. It always has lower speed and satisfying precision. Fair-MOT gives the\cite{zhang2021fairmot} propose the network that joint the re-ID and detection which share the fair feature. While the  Chained-Tracker\cite{peng2020chained} proposes an end-to-end model using adjacent frame pair as input and generating the box pair representing the same target. While \cite{SMILEtrack} introduced separate this into two moudles and gives similarity learning and tracking.
\paragraph{End to end}
 The end-to-end framework developed quickly with the successful proposal of transformer and its encoder and decoder framework. When a vision transformer is successfully applied to image recognition, transformer-based multi-object tracking is widely discussed. solved MOT based on end-to-end transformer\cite{zhu2021looking}. TransTrack\cite{sun2020transtrack} build on Deformable DETR with two decoders, solving this problem by matching the detection box between the two decoders. While the TransCenter is a point-based tracking that proposes a dense query feature map with a multi-scale of the input image for MOT leveraging transformer. GTR took short clip as input, encode object features and associated objects across all frames using global queries\cite{zhou2022global}, MeMOT\cite{cai2022memot} used a transformer to encode key information to large spatiotemporal memory, decode for solving detection and also the association.  MOTRv2\cite{zhang2022motrv2} using the YOLOX generates proposal queries and tracking by the MOTR\cite{zeng2022motr} self-attention scheme.
   
 Detectors in  MOT systems that use an independent frame often fail to fully utilize temporal information. To address this issue, end-to-end frameworks have been rapidly developed because of the successful proposition of transformer-based encoder and decoder frameworks.
 Transformer-based MOT approaches include TransTrack\cite{sun2020transtrack}, which builds on Deformable DETR with two decoders, allowing for detection box matching between them.  is TransCenter\cite{xu2021transcenter}, which proposes a dense query feature map with a multi-scale of the input image for MOT, leveraging transformers. GTR \cite{zhou2022global}takes short clips as input, encoding object features and associating objects across all frames using global queries. MeMOT\cite{cai2022memot}uses a transformer to encode key information into large spatio-temporal memory, decoding for both detection and association. Finally, MOTRv2\cite{zhang2022motrv2}  generates proposal queries and performs tracking using the MOTR self-attention scheme, based on the YOLOX model.
 Overall, these recent transformer-based MOT approaches represent significant progress in the field, potentially leading to more accurate and efficient tracking of multiple objects in complex environments.

\section{Mesh Tracker}
In this section, we present our three main modifications and improvements for the multi-object tracking-based
tracking-by-detection methods. The overall architecture is shown in the Fig. \ref{fig:architecture}. The pseudo-code of the proposed algorithm is in the appendix.
\begin{figure}[H]
	\centering  

    \includegraphics[width=0.92\textwidth]{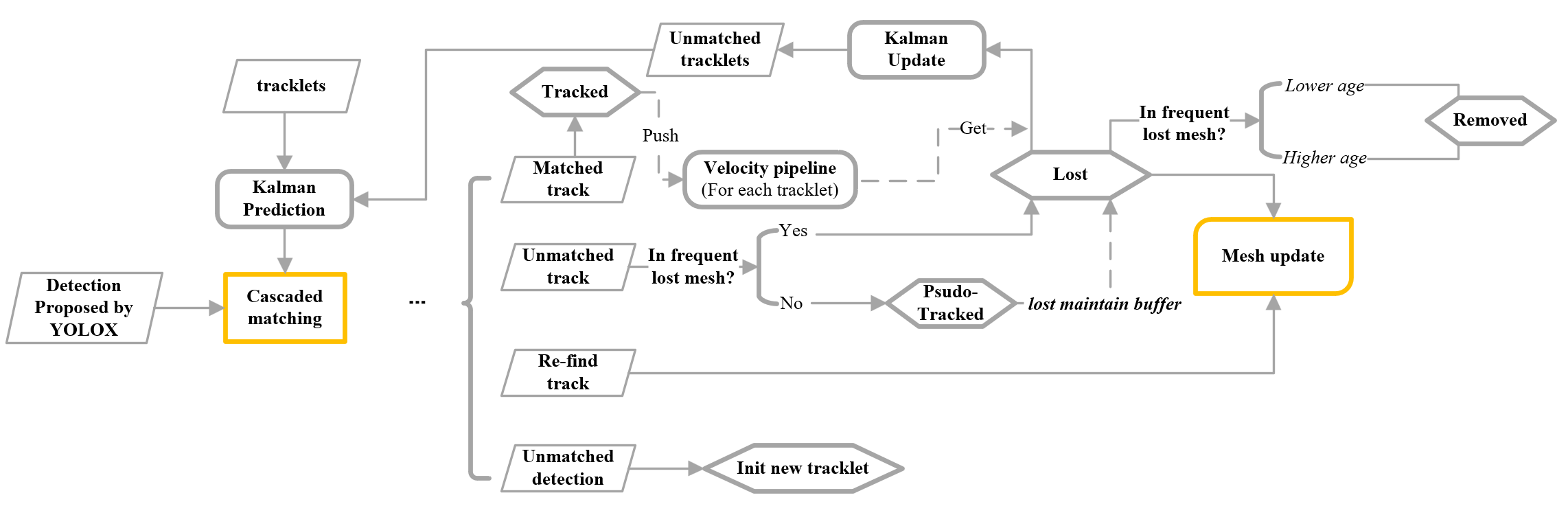}
    \caption{Framework of \textbf{Mesh-SORT}. It can be seen that it mainly deals with the lost tracklets management and many parameters can be turned adaptively (i.e. \textit{vel buffer, lost maintain buffer, location-wise ages}) based on the scene that needs to be tracked.}
    \label{fig:architecture}
    
\end{figure}

\subsection{Frequent Loss Region Identification}
  In some static camera motion scenes, the entry and exit points of the object into the picture are often recognizable. These points always correspond to real entry and exit. For fixed images with wide views, it is a good way to recognize them and applied different tracking strategies, due to the fact that if one object is out of frame, there is no need to track it.
On another side, the camera motion is stable, or in some specific scene (i.e. car, pedestrians on the street), the area of the object lost in most scenes typically has a certain degree of regularity over a given period.  We want to address this problem by dividing the frame space into independent sub-region and observing the special prior on this sub-region to identify them for applying different strategies. 

In this work, we defined the mesh as the equal division of frame space, and we set the number of objects temporal lost in the sub-region as our metrics to evaluate the frequent region because it can act as an indicator to represent that in some regions, the probability of an object being suddenly occluded is less than the probability of an object out of frame or occluded by fixed obstacles.  So we defined this count as:
\begin{equation}
    c_{i,j}= L_{i,j}-F_{i,j}
    \label{f1}
\end{equation}
Where $i,j$ are the horizontal and vertical mesh id respectively, $L_{i,j}$ is the number of objects lost in this mesh. And $F_{i,j}$ is the number of objects found back in this mesh. 

 After obtaining each mesh count, we can threshold them and identify the frequent mesh. Without loss of generality, to tackle the scene the number of tracks relatively remained constant and no obvious entry (i.e. dancing scenarios), the threshold can be reversed and heuristic functions can be made practical. Hence we defined the set of function space $\mathcal{H}$, and the threshold can be made from the set:
 \begin{equation}
    \begin{aligned}
    \mathcal{H}&=\{ h(s_{i,j},t) \mid t>0 \}\\
    \text{where}  \quad  s_{i,j}& =\left\{\begin{array}{cc}
        1&,m_{i,j}\text{ is frequent lost mesh} \\  
        0&,m_{i,j}\text{ is no frequent lost mesh}
        \end{array}  \right.
\end{aligned}
 \end{equation}
 Where $h(\cdot,\cdot)$ represents the threshold function, and $s$ is the state of mesh, this states that the threshold can also be adaptive when the state is changed or time-related. The Algorithm.\ref{alg:1} shows this simple algorithm for frequent mesh identification.
 \begin{algorithm}
	\renewcommand{\algorithmicrequire}{\textbf{Input:}}
	\renewcommand{\algorithmicensure}{\textbf{Output:}}
	\caption{Simple mesh identification algorithm}
	\label{alg:1}
	\begin{algorithmic}[1]
	\REQUIRE horizontal segmentation $m$, vertical segmentation $n$ ; identification threshold $h(s,t)$
	\ENSURE  Frequent lost Mesh on image region $\bf{M}=\{m_1,m_2,\dots\}$;
	\STATE Construct the $m \times n$ mesh matrix
    \FOR{each lost tracklet}
        \STATE Compute the  center bottom point $(x_{c},y_{b})$ of bounding box
        \STATE Find the correspond mesh $m_{i,j}$
        \STATE Count $c_{i,j}=c_{i,j}+1$
    \ENDFOR
    \FOR{each re-find tracklet}
        \STATE Compute the center bottom point $(x_{c},y_{b})$ of bounding box
        \STATE Find the correspond mesh $m_{i,j}$
        \STATE Count $c_{i,j}=c_{i,j}-1$
    \ENDFOR
    \FOR{$i$ in $m$}
    \FOR{$j$ in $n$}
    \IF{ $c_{i,j}>h(s,t)$}
   \STATE  Add $m_{i,j}$ into $\bf{M}$
        \ELSE
       \STATE Remove  $m_{i,j}$ into $\bf{M}$
        \ENDIF
    \ENDFOR
    \ENDFOR
	\end{algorithmic}  
\end{algorithm}



\begin{figure}[h]
	\centering  
	\subfigure[Frequent loss Mesh ID: \{(0,0)\}
    ]{
		\includegraphics[height=0.3\linewidth]{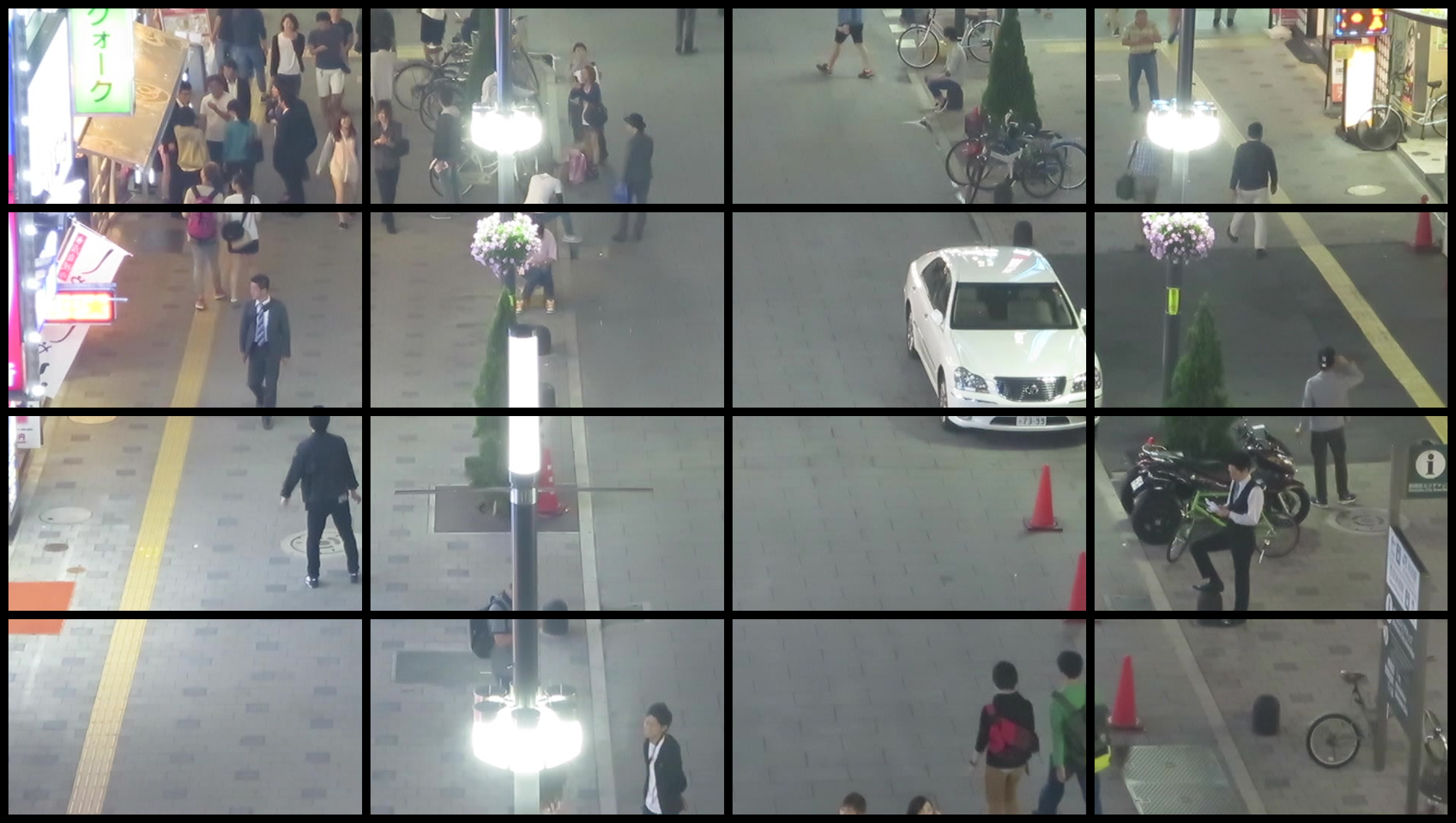}}
	\subfigure[Correspond heatmap]{
		\includegraphics[height=0.25\linewidth]{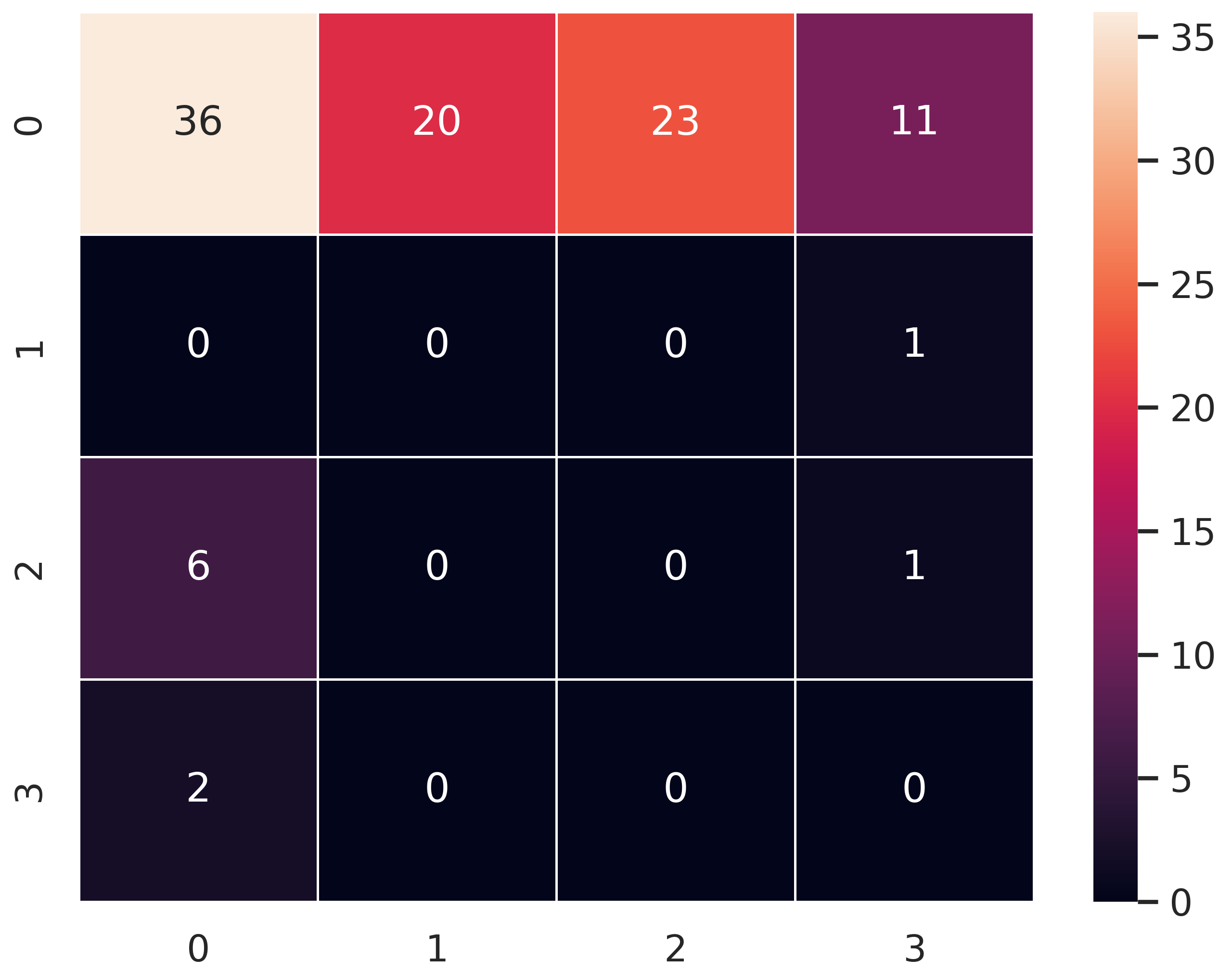}}
	\caption{Visual illustrations of mesh division and  frequent loss mesh identification  in MOT17 datasets(\textit{Frame ID: 680})}
    \label{fig:mesh}
\end{figure}

 Fig. \ref{fig:mesh} shows the example of the mesh identification and the count of lost computed by eq.\eqref{f1}. We can obtain from the figure that it can simply reveal the density of the tracking object and make the sub-region independent of others for applying different strategies. By introducing non-maximum suppression (NMS), it can also be a possible solution for estimating the potential obstacle and relatively crowded scenes.
\subsubsection{Location-wise Lost Age}

\begin{figure}[hbtp]
	\centering  
		\includegraphics[width=0.7\linewidth]{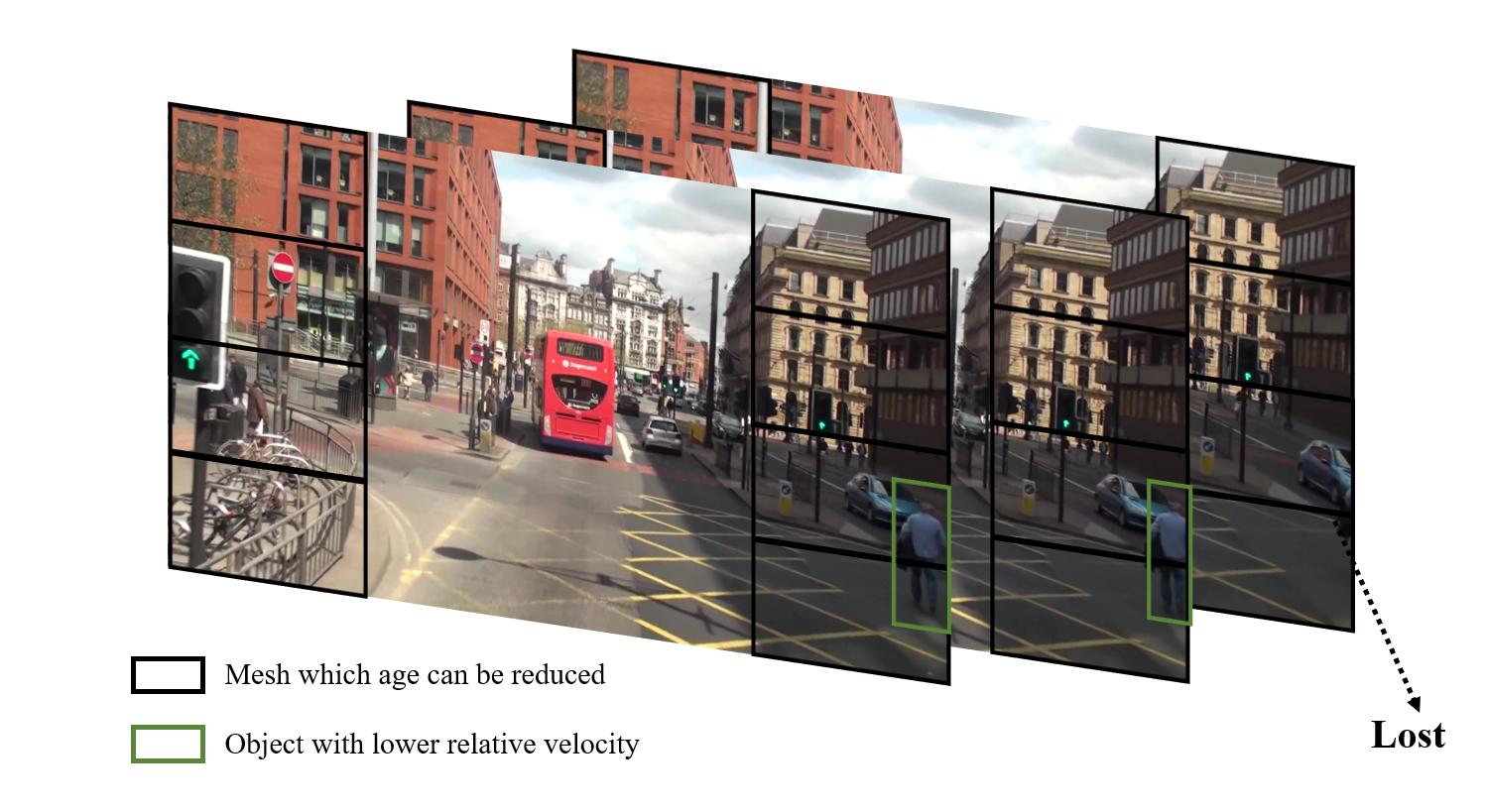}
        \caption{ Camera motion with fixed objects passing sub-region in dataset MOT17-14, in this figure, the person with green box will not be expected tracked because the camera with relatively higher velocity account to the person}
    \label{fig:motion}
\end{figure}

After identifying frequent loss sub-regions, a change in the policy of lost management may be necessary. Loss buffers can be adaptive because different sub-regions may lead to different occlusion situations. 
For example, it is a high possibility that frequently lost regions represent exit or entry in real-world locations. If an object is observed to be out of frame, it is no longer expected to be tracked, and its loss buffer (tracking ages) can be reduced.  
For the situation with camera motion, the camera will pass objects with relatively low velocity, these will also not be expected for tracking. Fig. \ref{fig:motion} demonstrates this situation, the camera will pass the objects, and the objects will out of frame.

It is believed that changing the age of objects in this region can be adaptive and reduce computational costs when mesh size is suitable, so we propose dual ages, which can be named as location wise ages module, if an object is lost in frequent loss sub-region, the ages can be reduced for optimize this sub-region for tracking. Also, it can also optimize the memory of devices and might decrease the situation of that wrong ID-assigned after frequent occlusion and overlap when the original object was re-initialized and the old tracklet will keep matching in the following period.


\subsection{Lost Tracklet Managment}

\subsubsection{Location-wise Lost Maintain}
\begin{figure}[h]
    
	\centering  
   \textbf{Without LM}\\
	\subfigure[tracked]{
		\includegraphics[width=0.2\linewidth]{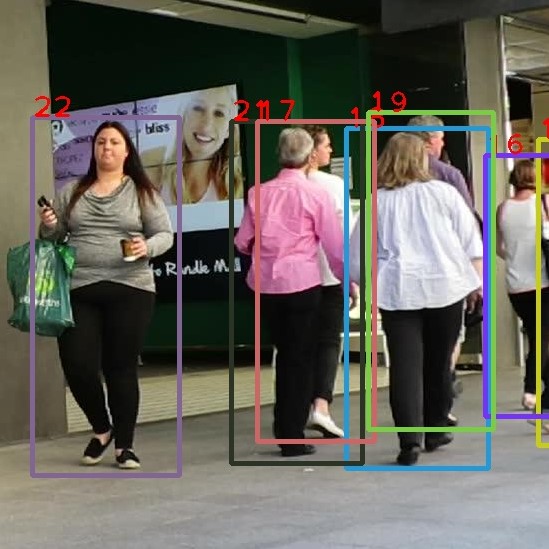}}
	\subfigure[lost]{
		\includegraphics[width=0.2\linewidth]{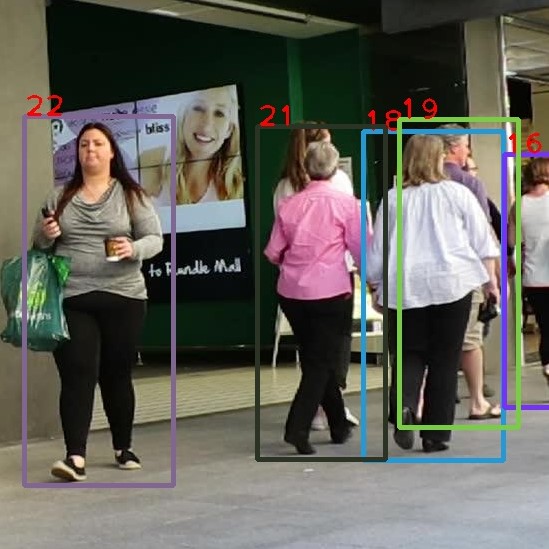}}
	\subfigure[Wrong assigned]{
		\includegraphics[width=0.2\linewidth]{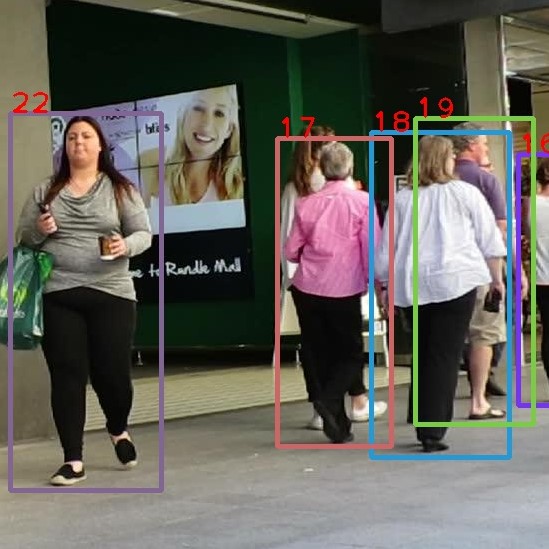}}
   \\ \textbf{Applying LM}\\
		\includegraphics[width=0.2\linewidth]{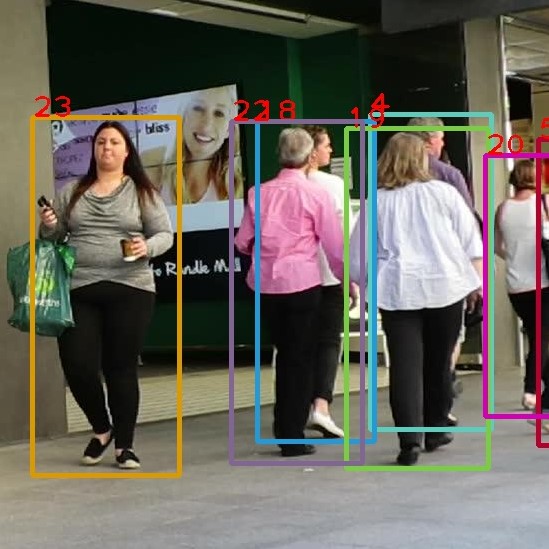}
		\includegraphics[width=0.2\linewidth]{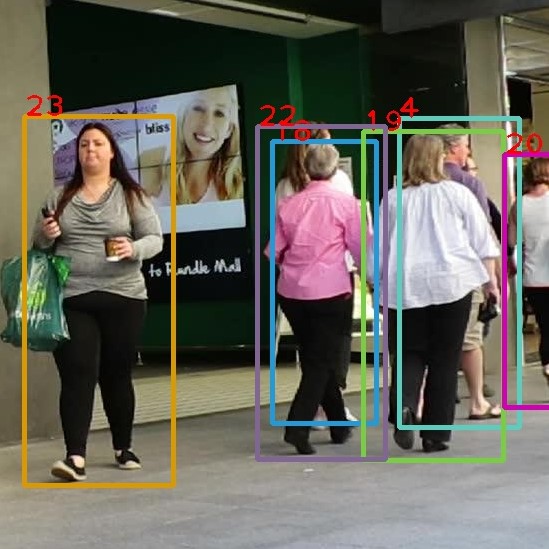}
		\includegraphics[width=0.2\linewidth]{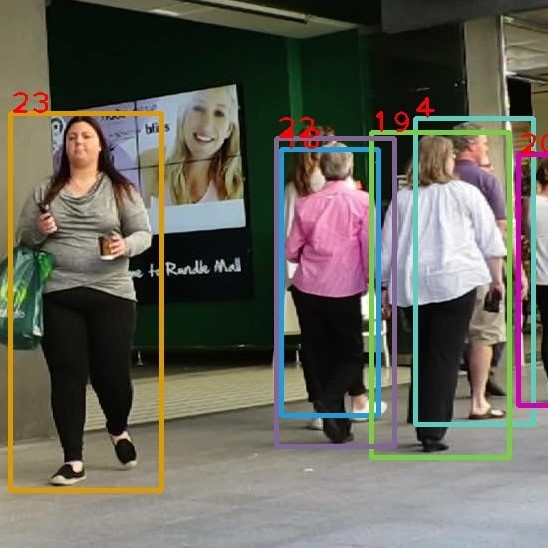}
	\caption{Comparison about Lost maintain under the situation of wrong ID assigned after occlusion in the MOT17 training datasets \textit{(Without LM ID:17, After applying ID:18) }}
    \label{fig:idsw}
\end{figure}

If the objects are not in high-frequent loss sub-regions, mesh trackers can identify them as occlusions or undetectable by the detector. To address this issue, we propose the Lost Maintain (LM) mechanism to maintain tracks with a certain number of frames. Suppose the lost maintain buffer $l$, the object can be put into a lost pool after continuous $l$ frames unmatched.  This mechanism allows the object to update its Kalman filter parameters and provides a virtual proposal of the tracklet.
Even though the lost tracks are still updated in the Kalman filter, the following matching also will be affected especially when two objects are overlapping.
In other words, this mechanism allows for more robust handling of tracker loss,  sudden failure of the detector or noise due to partial occlusion can be re-assigned.   As Fig.\ref{fig:idsw}  shows, this mechanism can find some occlusions back without lost and it can also increase the consistency of tracking and reduce fragmentation caused by a bad detector in the case of temporary occlusions.

\subsection{Detector Error Magnification}

\begin{figure}[hbtp!]
	\centering  
	\subfigure[Original]{
		\includegraphics[width=0.2\linewidth]{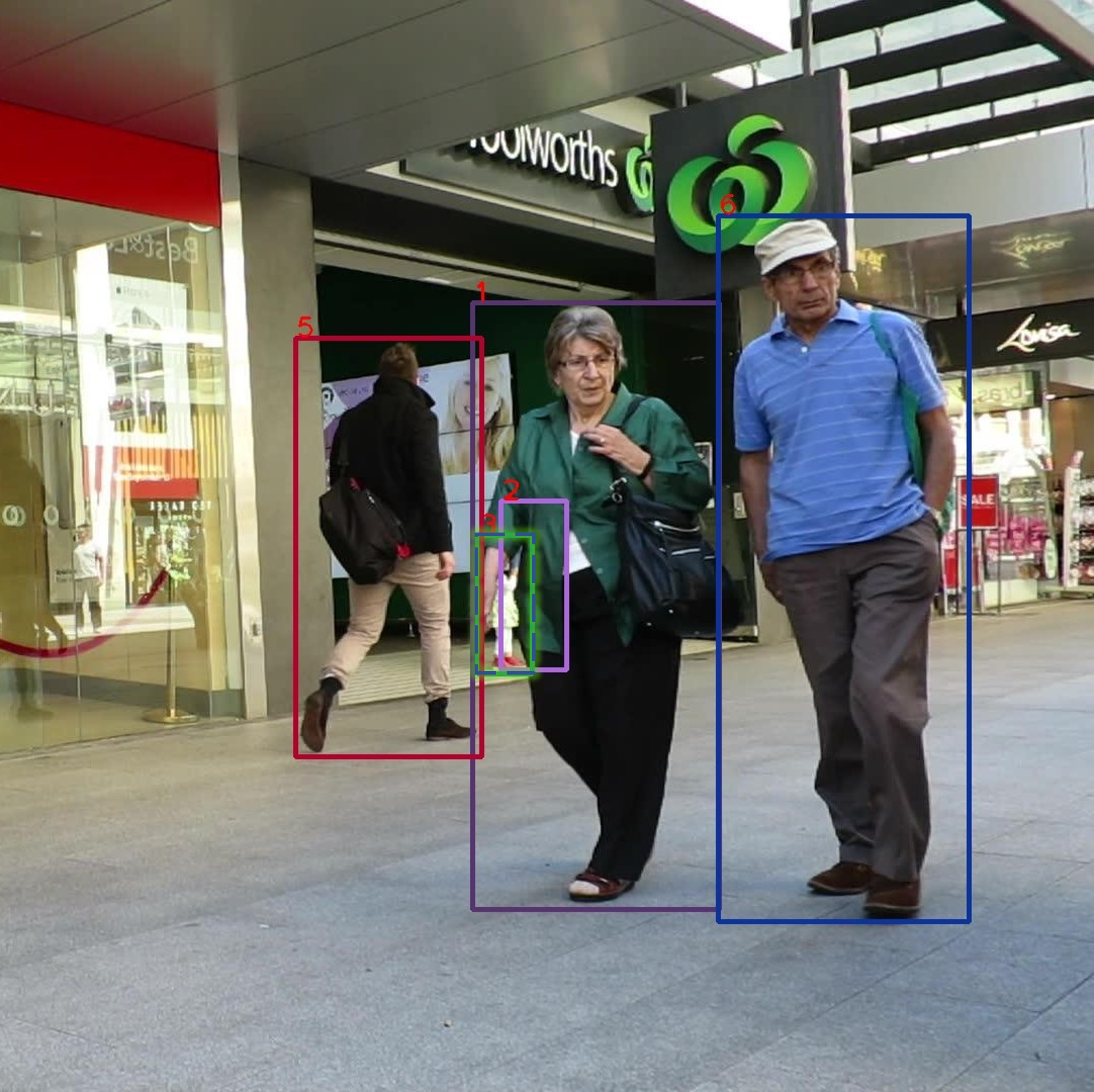}}
	\subfigure[Semi-occluded ]{
		\includegraphics[width=0.2\linewidth]{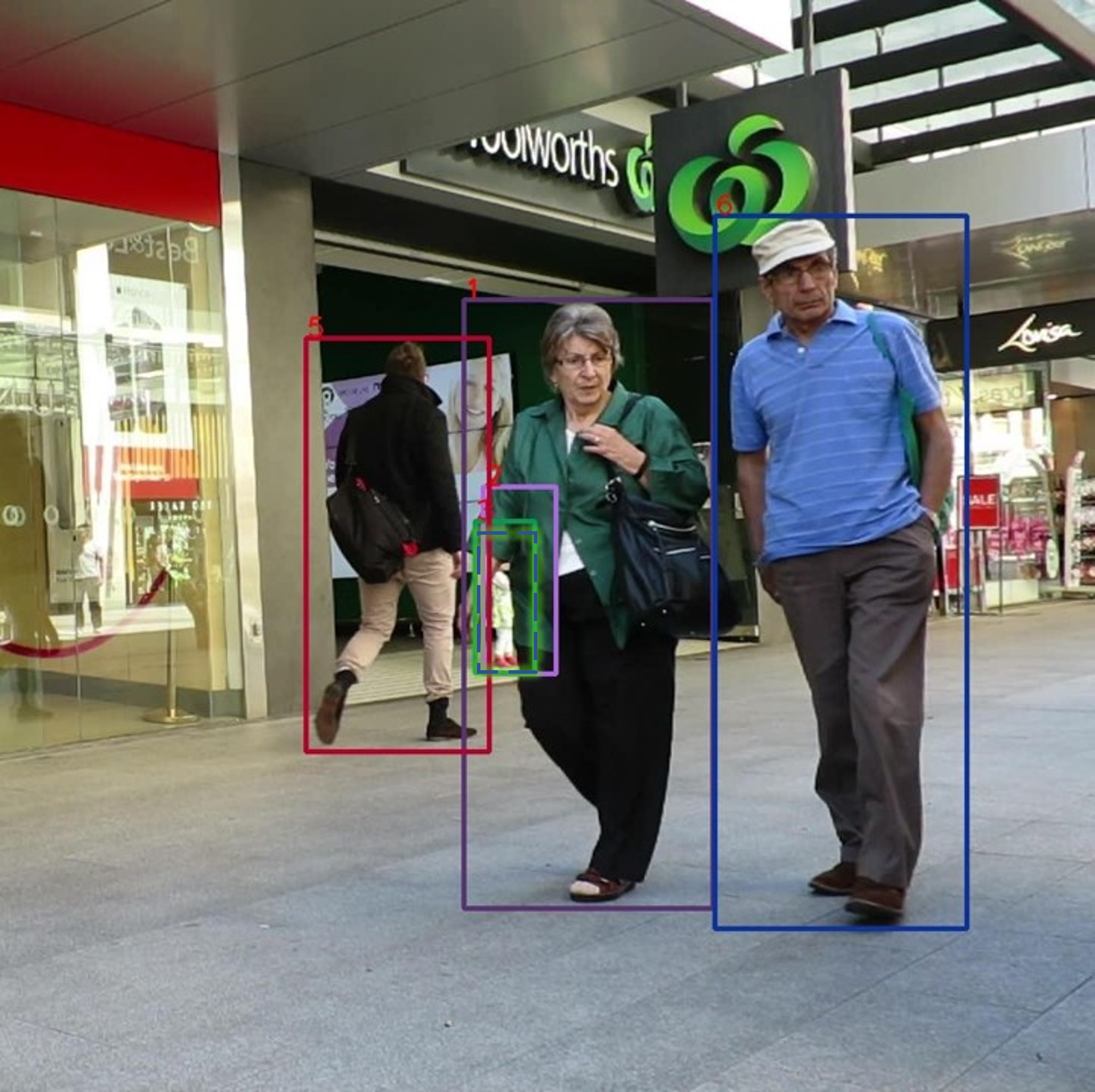}}
	\subfigure[Occluded]{
		\includegraphics[width=0.2\linewidth]{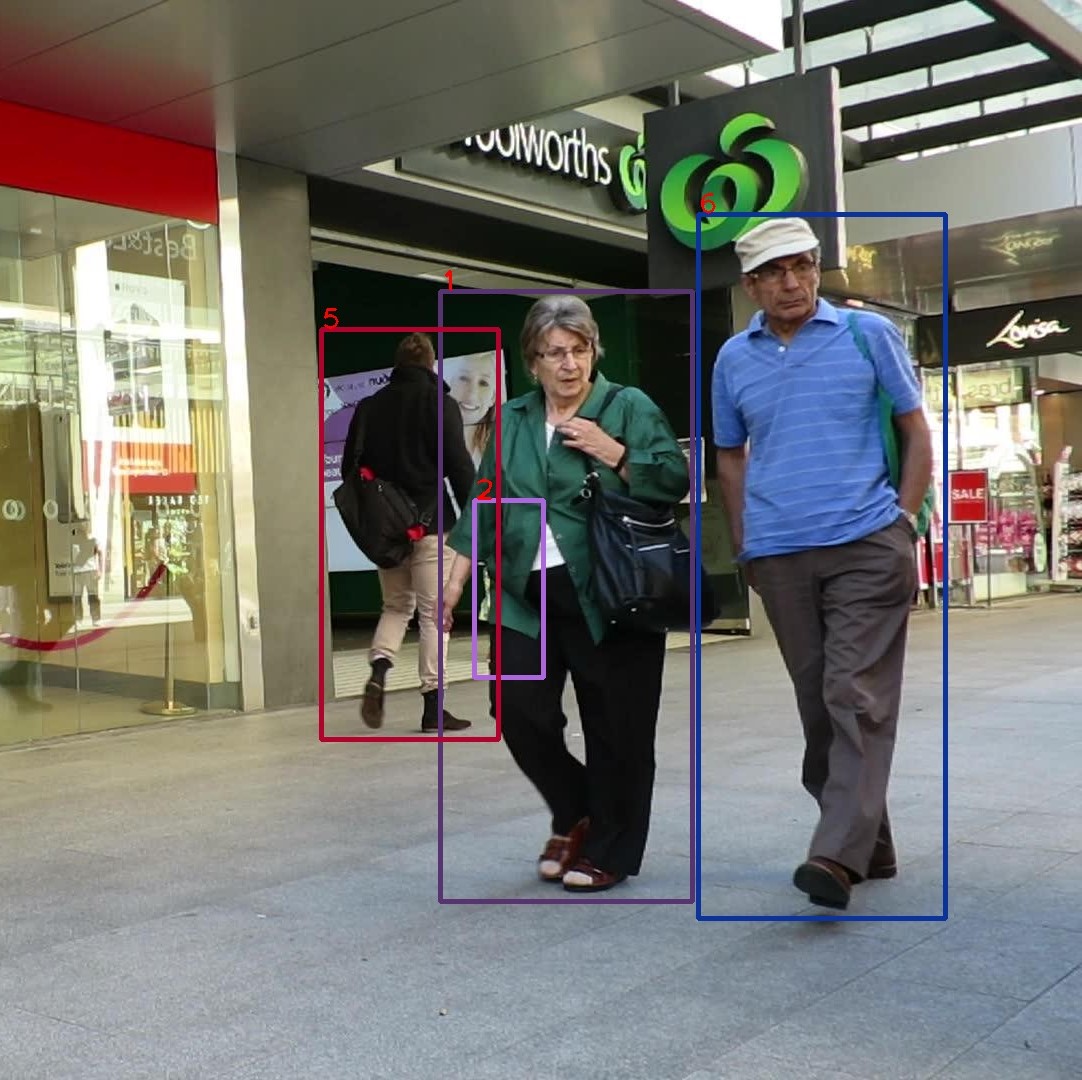}}
	\caption{bounding boxes scale change  under semi-occluded situation (\textit{Track ID: 3}) }
    \label{fig:vc}
\end{figure}

 When the occlusion happens, the detection may have a sharp size change. Fig.\ref{fig:vc} demonstrates the bounding box scale change under the semi-occluded situation. This error may be captured by the Kalman filter, and magnified during the occlusion. Inspired by OC-SORT \cite{cao2022ocsort} ,when
no observations are provided to the Kalman filter, it will recursively update, variance of state estimates will be proportional to the time. 

Suppose one object's Kalman state $\bm{x}_t=[x_c,y_c,a,h, \dot{x_c},\dot{y_c},\dot{a},\dot{h}]^T \in \mathbb{R}^{8}$ at time $t$, where $(x_c,y_c)$ are coordinates of the object center in the image plane, $a$ is the bounding box scale and $h$ is the bounding box aspect ratio; and observation at time $t+1$ is $z_{t}=[x_c^\prime,y_c^\prime,a^\prime,h^\prime]^T \in \mathbb{R}^{4}$; bad detector proposal can be modeled as the noise $\epsilon_a \sim \mathcal{N}(0,\sigma_a^2), \epsilon_h \sim \mathcal{N}(0,\sigma_h^2)$ performed on bounding box scale and ratio. 

In the next frame with time $t+1$,
if the semi-occluded happens on this object, the Kalman filter updating rules after matching given by (Appendix \ref{kalman}):
    \begin{equation}
        \begin{aligned}
        \bm{x}_{t+1} &= \hat{\bm{x}}_{t} +K_t((z_{t}+\sigma)-H_t\hat{x}_{k-1})\\
        &= \hat{\bm{x}}_{t} +K_t(z_{t}-H_t\bm{\hat{x}}_{t})+K_t\sigma
    \end{aligned}
    \end{equation}
    where $\sigma$ defined as:
\begin{equation}
   \sigma= [0,0,\epsilon_a,\epsilon_h]^T
\end{equation}
In the following $k$ frame, the occlusion happened and no observation was provided to KF, the noise will keep iter and prediction (Appendix \ref{Kal:est}):
\begin{equation}
    \begin{aligned}
    \bm{\hat{x}}_{t+k} &= F^{k-1} \hat{\bm{x}}_{t+1} \\
     &=  F^{k-1} (\hat{\bm{x}}_{t} +K_t(z_{t-1}-H_t\bm{\hat{x}}_{k-1}))+F^{k-1}K_t\sigma
\end{aligned}
\end{equation}
It demonstrates the effect that a bad proposal of a semi-occluded object in the previous frame can have on the prediction. To mitigate this effect, we propose a \textbf{velocity buffer}, which can store a certain amount of previous velocity in a tracklet, when the occlusion happened, the current velocity in the Kalman filter will be replaced by the previous velocity in the Kalman Filter, to provide a more robust Kalman prediction. Because in the linear motion model, the velocity of object will not vary strongly in a few frames, so a replacement can not be used with a long buffer.  After applying the velocity buffer, the prediction will be more accurate and noise can be reduced, the Fig.\ref{fig:vc} demonstrates the situation that Finding switched ID back after overlap and non-linear motion on the Dancetrack.

\begin{figure}[hbtp!]
   
	\centering  
    \textbf{Without applying vel-buffer}\\
	\subfigure[Original]{
		\includegraphics[width=0.2\linewidth]{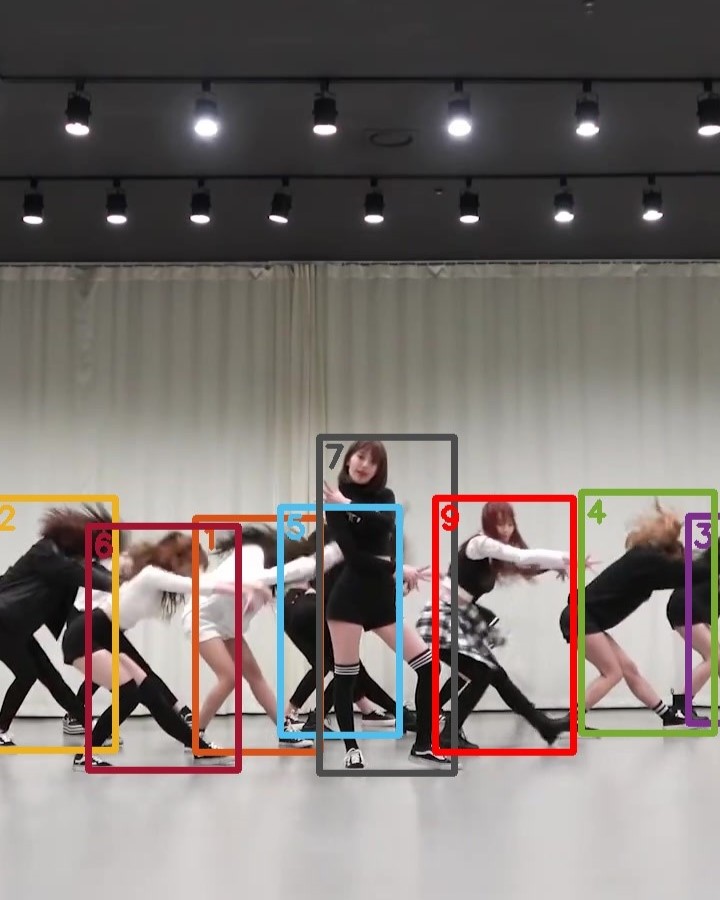}}
	\subfigure[Overlap]{
		\includegraphics[width=0.2\linewidth]{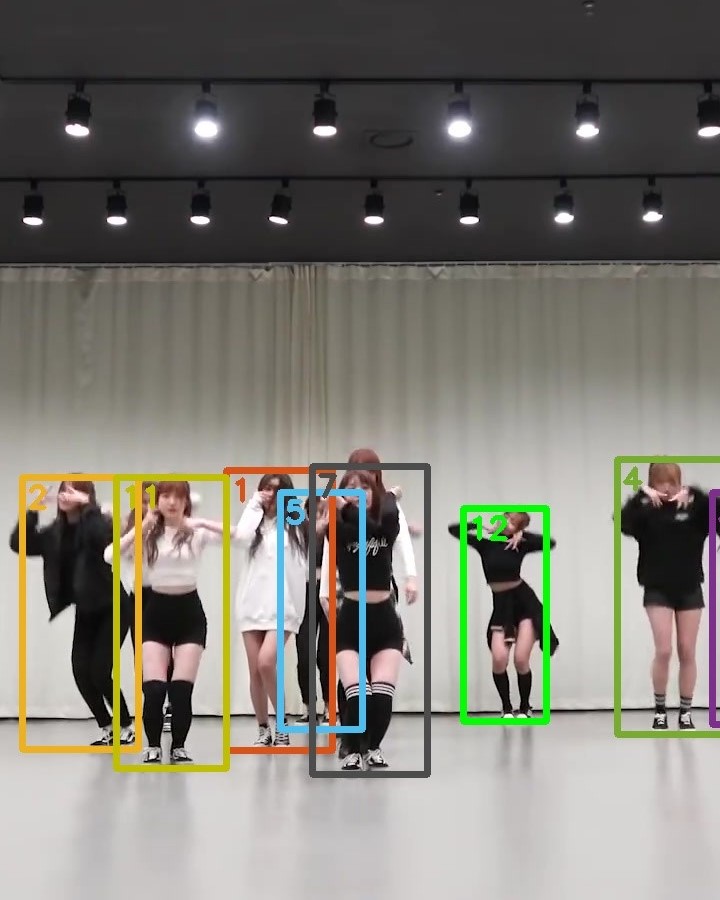}}
	\subfigure[ID switches]{
		\includegraphics[width=0.2\linewidth]{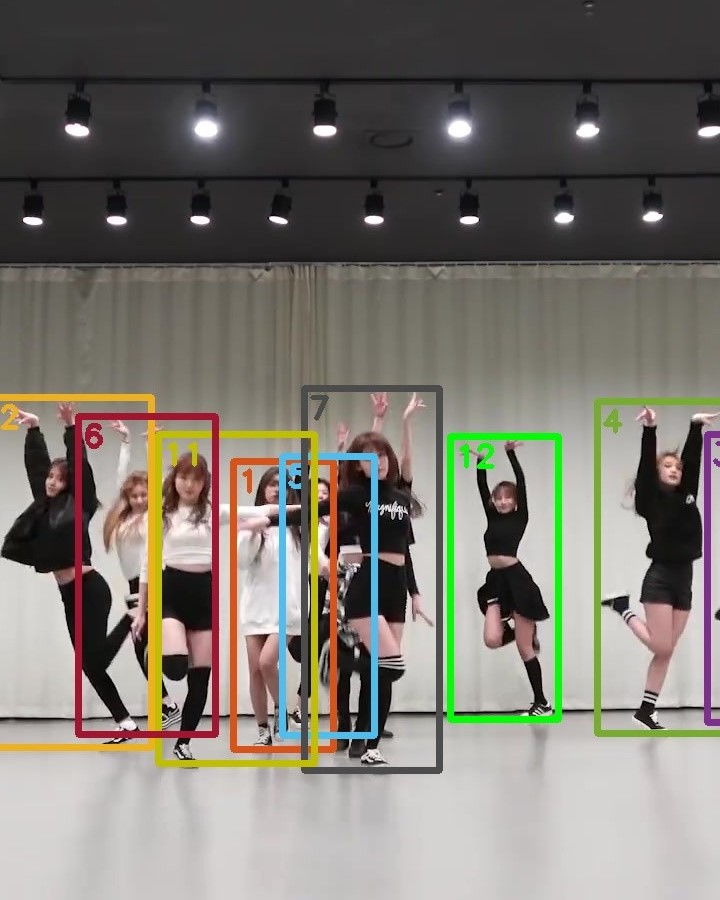}}
   \\ \textbf{Applying vel-buffer}\\
		\includegraphics[width=0.2\linewidth]{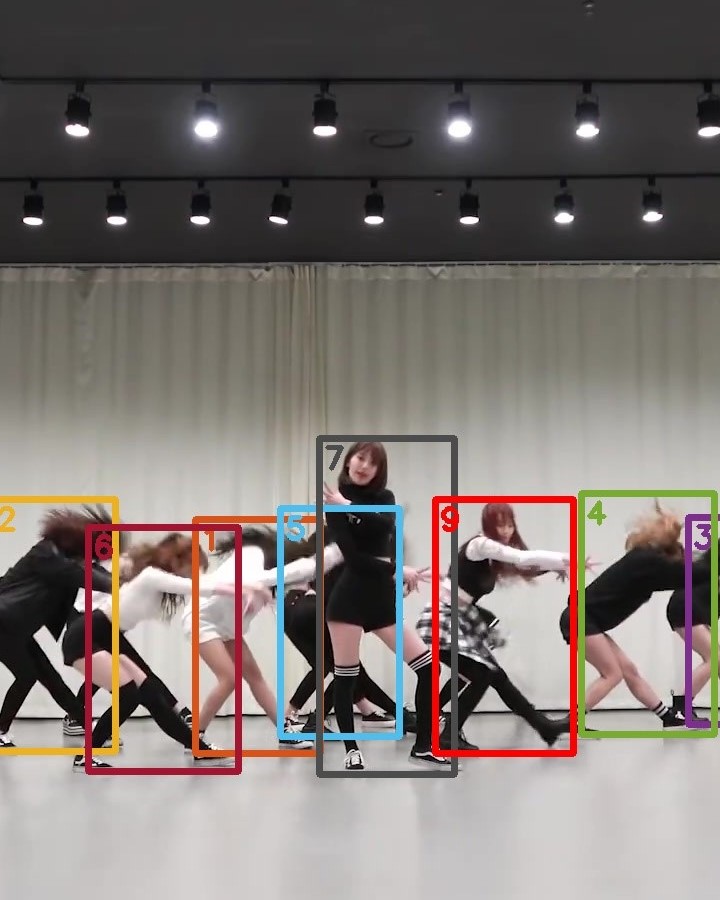}
		\includegraphics[width=0.2\linewidth]{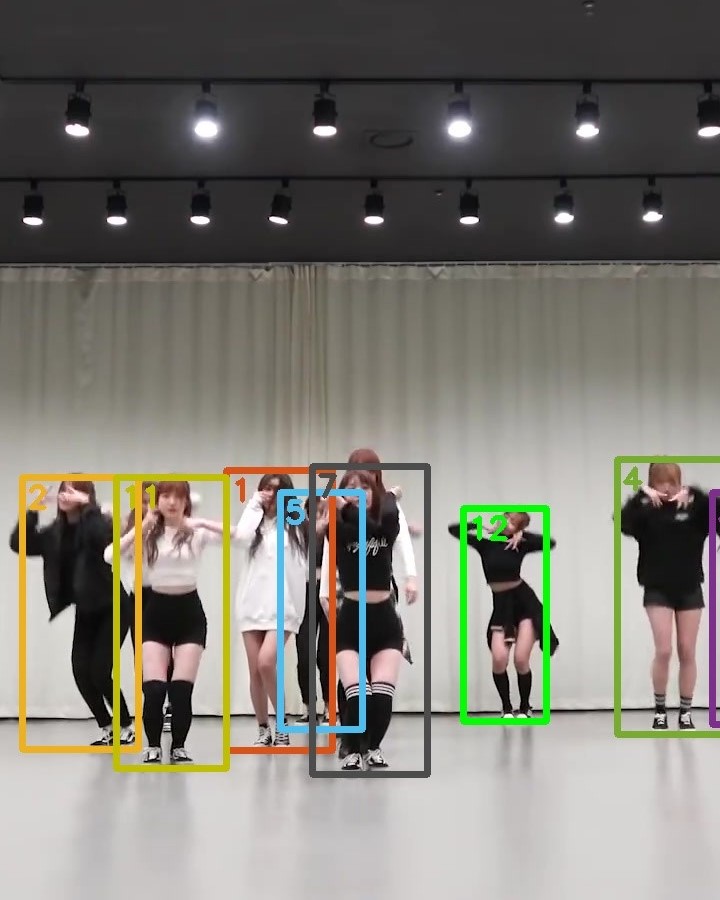}
		\includegraphics[width=0.2\linewidth]{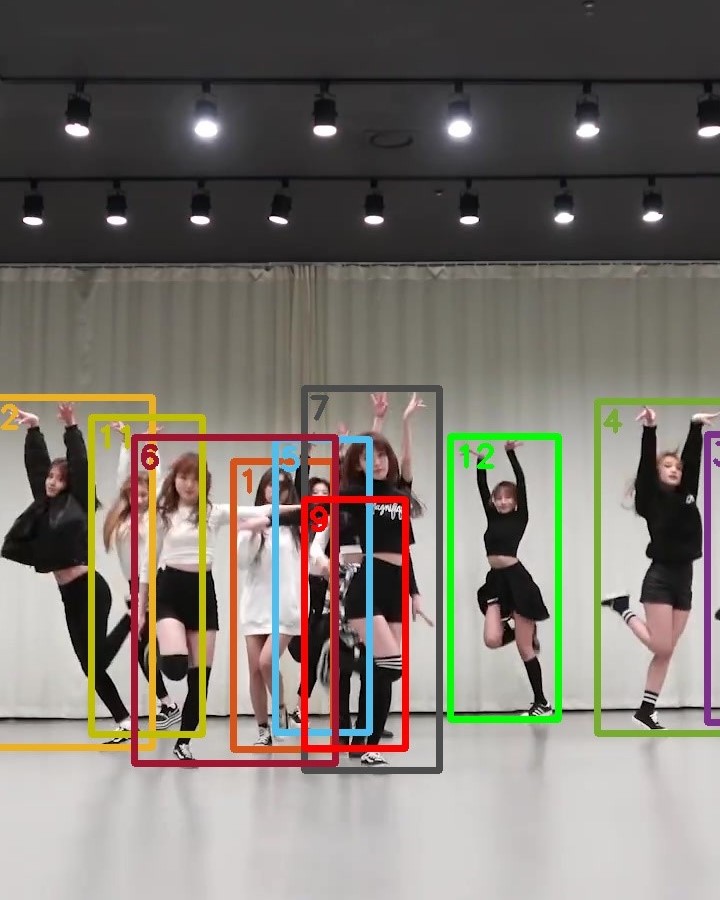}
	\caption{Comparison of application of velocity buffer for finding back ID after overlapping on dancetrack datasets (\textit{Track ID:6,11})}
    \label{fig:vl}
\end{figure}


\section{Experiments}
In this section, we provide experiments to demonstrate
the efficiency of Mesh-SORT on multiple datasets. 
\subsection{Experiments settings}
\paragraph{Datasets.} In our experiment,  MOT17, MOT20\cite{milan2016mot16} has been introduced for verify our methods. They are the most popular benchmark of multi-object tracking with pedestrian scenes, and their motion is mostly near linear.  And Dancetrack\cite{peize2021dance} also proposed in recent years become a popular dataset, it emphasizes the association part of the algorithm. Because dancing objects localization is easy but the objection motion is non-linear in this dataset. We choose datasets containing two scenes, which can help us make a comprehensive analysis.

\paragraph{Metrics.} We use MOTA, Identity F1 Score (IDF1), ID switches(IDs), fragmentation(FM), Mostly Tracked(ML), HOTA\cite{luiten2021hota} for our experiment. Among them, HOTA and MOTA are the main metrics, where classical performance index MOTA is computed using the FP (False Positive), FN (False Negative), IDs (Id switches), while HOTA\cite{luiten2021hota} is a metrics-defined performance combined with detection association and localization.  Considering the tracking inference speed, the frame per second (FPS)  as a relative speed indicator will be compared in experiments.

The formula of MOTA and IDF1 is shown on equation \ref{MOTA} and \ref{IDF1}, the formula of HOTA is presented in equation \ref{hota}
\begin{equation}
\label{MOTA}
M O T A=1-\frac{\sum_t\left(F N_t+F P_t+I D S W_t\right)}{\sum_t G T_t}
\end{equation}
\begin{equation}
\label{IDF1}
I D F_1=\frac{2 I D T P}{2 I D T P+I D F P+I D F N}
\end{equation}
\begin{equation}
\label{hota}
\mathrm{HOTA}_\alpha=\sqrt{\frac{\sum_{c \in\{\mathrm{TP}\}} \mathcal{A}(c)}{|\mathrm{TP}|+|\mathrm{FN}|+|\mathrm{FP}|}}
\end{equation}
where $\mathcal{A}(c)$ is 
 association accuracy score, which measures the alignment of the trajectory of ground truth. 
 
\textbf{Implementation settings.} Even though there are other state of art algorithms, we still select bytetrack\cite{zhang2021bytetrack} as our baseline, which break the detection into higher confidence bounding boxes and lower bounding boxes and matches separately. For a fair comparison, we use the publicly available detector YOLOX\cite{ge2021yolox}. Our method runs on a device with Intel(R) Xeon(R) CPU E5-2680 v3 @ 2.50GHz and two GTX 3090 GPUs as the benchmark of FPS.

\textbf{Implementation.} For algorithm implementation, we implemented the Mesh using bound boxes location simply division by mesh size;
\begin{equation}
    i=\frac{x_{bm}}{m} \quad  j=\frac{y_{bm}}{n}
\end{equation}
Where $m, n$ is horizontal and vertical mesh segmentation respectively, $x_{bm},y_{bm}$ are the bottom middle of bounding boxes and the $i,j$ is mesh id in a 2-D mesh matrix.  For the threshold function, we choose the following linear time-variant function for keeping simplicity in all datasets.
\begin{equation}
    \begin{aligned} h(s_{i,j},t)= \left\{
    \begin{array}{cc}  
       \lambda t,& s=0\\ 
                        0,& s=1
    \end{array}
    \right.
\end{aligned}
\end{equation}

Where $\lambda$ can be adjust based on the mesh size. Meanwhile, we implemented a velocity buffer by queue, the selection of the buffer will be discussed in the next section.

\subsection{Ablation Study}
\paragraph{Overall} To demonstrate the effectiveness of the proposed modules, we perform an ablation study on validation sets of MOT17.  The component's vel-buffer and lost maintain mechanism and location-wise buffer.  The results are shown in Table \ref{ablation}. We found that after applying these three modules, MOTA and IDF1 were improved and the IDs were significantly reduced.

\paragraph{Benefit of Lost maintain} The table.\ref{table:lm} shows different buffers of lost maintain mechanism act on the tracker caused the change of Fragmentation (FM) and Mostly Tracked (MT) in MOT17 datasets. From the data it can be found that it can keep MOTA stable if the buffer is within a suitable range.  And the performance of fragmentation and mostly tracked  was significantly improved, which demonstrate the theory that it can improve tracking consistency and dealing with bad detector proposals.

\paragraph{Effectiveness of Mesh} The choice of mesh size was discussed in table.\ref{table:mesh}, where the threhold are set to be $\lambda=0.02$, and location-wise age $8$ lower than the normal ages.  It can obtain from data that mesh identification and location-wise age can independently improve FPS without losing the MOTA and other performance.

We find that if we apply these modules' parameter without tuning, the result on the Dancetrack show little improvement because the scene is paid more attention to the association, which allow us to carefully turned the threshold function to make the result more adaptive to the scene.

\begin{table*}[hbtp!]
    \begin{center}
    \scalebox{0.95}{\begin{tabular}{c |c c c}
    \textbf{Method}
     & \textbf{MOTA} $\uparrow$& \textbf{FM}$\downarrow$ &\textbf{Most Tracked}$\downarrow$  \\
         \toprule 
     Baseline            &        76.5 & 1411& 594 \\
     Baseline+Lost Matain($l=1$)  &  76.5  & 1334  & 614  \\
     Baseline+Lost Matain($l=3$)  &   76.5 &1250 & 633  \\
     Baseline+Lost Matain($l=5$)  &  76.2 & 1213 &635 \\    
  \bottomrule
 \end{tabular}}
    \end{center}
    \vspace{-2mm}
    \caption{Ablation study of  lost maintain mechanism of Mesh-SORT on MOT17 training set under the pretrained YOLOX detector.}
    \label{table:lm}
    \end{table*}    
    \setlength{\tabcolsep}{10.0pt}

\begin{table*}[hbtp!]
    \begin{center}
    \scalebox{0.95}{\begin{tabular}{c |c c c}
    \textbf{Method}
     & \textbf{MOTA} $\uparrow$& \textbf{MT}$\uparrow$ &\textbf{FPS}$\uparrow$ \\
         \toprule 
     Baseline            &     76.5    &     594           & 15.4  \\

     Baseline+ $3\times3$ Mesh,8 lower age  & 76.5  & 605  &  15.7    \\
     Baseline+ $4\times4$ Mesh,8 lower ages & 76.6  & 612  &  15.6  \\
     Baseline+ $5\times5$ Mesh,8 lower ages &  76.6 & 605  & 15.6\\    
     Baseline+ $6\times6$ Mesh,8 lower ages & 76.6  &  602 & 15.2\\    
  \bottomrule
 \end{tabular}}
    \end{center}
    \vspace{-2mm}
    \caption{Ablation study of  mesh size of Mesh-SORT on MOT17 training set under the pretrained YOLOX detector.}
    \label{table:mesh}
    \end{table*}    
    \setlength{\tabcolsep}{10.0pt}

\begin{table*}[hbtp!]
    \begin{center}
    \scalebox{0.95}{\begin{tabular}{c |c c c| c c c}
    \textbf{Method}  & \textbf{Lost Maintain} & \textbf{Vel-buffer}& \textbf{Location-wise ages}
     & \textbf{MOTA} & \textbf{IDF1} &\textbf{IDs} \\
         \toprule 
     Baseline (ByteTrack*) &-           &-& - & 76.5 &  79.4& 500 \\
     Baseline + Column1    & \checkmark &- & -  & 76.0  & 78.6  & 485  \\
     Baseline + Column2    &\checkmark  &\checkmark & - &  76.6 &78.6 & 476  \\
     Baseline + Column1-3  & \checkmark & \checkmark  & \checkmark & \textbf{76.9} & \textbf{79.9} &\textbf{464} \\    
  \bottomrule
 \end{tabular}}
    \end{center}
    \vspace{-2mm}
    \caption{Ablation study of Mesh-SORT on MOT17 test set under the pre-trained YOLOX detector. Note that the mesh identification strategies have been added to the lost maintain mechanism}
    \label{ablation}
    \end{table*}    
    \setlength{\tabcolsep}{10.0pt}

\subsection{Benchmark results}

\begin{table*}[hbtp!]
    \begin{center}
    \scalebox{0.95}{\setlength{\tabcolsep}{10.0pt}

\begin{tabular}{ l | c | c | c | c | c | c | c} 
   
\toprule
Tracker & MOTA$\uparrow$ & IDF1$\uparrow$ & HOTA$\uparrow$ & FP$\downarrow$ & FN$\downarrow$ & IDs$\downarrow$ & FPS$\uparrow$ \\
\midrule
MOTR \cite{zeng2021motr}                    & 65.1 & 66.4 & - & 45486 & 149307 & 2049 & -\\
CTracker \cite{peng2020chained}             & 66.6 & 57.4 & 49.0 & 22284 & 160491 & 5529 & 6.8\\
CenterTrack \cite{zhou2020tracking}         & 67.8 & 64.7 & 52.2 & 18498 & 160332 & 3039 & 17.5\\
QuasiDense \cite{pang2021quasi}             & 68.7 & 66.3 & 53.9 & 26589 & 146643 & 3378 & 20.3\\
TraDes \cite{wu2021track}                   & 69.1 & 63.9 & 52.7 & 20892 & 150060 & 3555 & 17.5\\
MAT \cite{han2022mat}                       & 69.5 & 63.1 & 53.8 & 30660 & 138741 & 2844 & 9.0\\

TransCenter \cite{xu2021transcenter}        & 73.2 & 62.2 & 54.5 & 23112 & 123738 & 4614 & 1.0\\
GSDT \cite{wang2020joint}                   & 73.2 & 66.5 & 55.2 & 26397 & 120666 & 3891 & 4.9\\
Semi-TCL \cite{li2021semi}                  & 73.3 & 73.2 & 59.8 & 22944 & 124980 & 2790 & -\\
FairMOT \cite{zhang2021fairmot}             & 73.7 & 72.3 & 59.3 & 27507 & 117477 & 3303 & 25.9\\
RelationTrack \cite{yu2021relationtrack}    & 73.8 & 74.7 & 61.0 & 27999 & 118623 & 1374 & 8.5\\
PermaTrackPr \cite{tokmakov2021learning}    & 73.8 & 68.9 & 55.5 & 28998 & 115104 & 3699 & 11.9\\
CSTrack \cite{liang2020rethinking}          & 74.9 & 72.6 & 59.3 & 23847 & 114303 & 3567 & 15.8\\
TransTrack \cite{sun2020transtrack}         & 75.2 & 63.5 & 54.1 & 50157 & 86442 & 3603 & 10.0\\
FUFET \cite{shan2020tracklets}              & 76.2 & 68.0 & 57.9 & 32796 & 98475 & 3237 & 6.8\\
CorrTracker \cite{wang2021multiple}         & 76.5 & 73.6 & 60.7 & 29808 & 99510 & 3369 & 15.6\\
TransMOT \cite{chu2021transmot}             & 76.7 & 75.1 & 61.7 & 36231 & 93150 & 2346 & 9.6\\
ReMOT \cite{yang2021remot}                  & 77.0 & 72.0 & 59.7 & 33204 & 93612 & 2853 & 1.8\\
MAATrack \cite{stadler2022modelling}        & 79.4 & 75.9 & 62.0 & 37320 & 77661 & 1452 & \textbf{189.1}\\
OCSORT \cite{cao2022observation}            & 78.0 & 77.5 & 63.2 & \textbf{15129} & 107055 & 1950 & 29.0\\
StrongSORT++ \cite{du2022strongsort}        & 79.6 & 79.5 & 64.4 & 27876 & 86205 & \textbf{1194} & 7.1\\ 
\midrule
ByteTrack(Baseline) \cite{zhang2021bytetrack}         & 80.3 & 77.3 & 63.1 & 25491 & 83721 & 2196 & 29.6 \\ 
\textbf{Mesh-SORT (ours)}                   & 80.4   &  78.0  &63.2  & 22023 & \textbf{81492} & 1923 & 29.8\\

\bottomrule
\end{tabular}

}
    \end{center}
    \vspace{-2mm}
    \caption{Comparison of the state-of-the-art methods under the “private detector” protocol on MOT17 test set. The best results are shown in \textbf{bold}.}
    \label{table_mot17}
\end{table*}    
    \setlength{\tabcolsep}{10.0pt}

    \begin{table*}[hbtp!]
        \begin{center}
        \scalebox{0.95}{
    \begin{tabular}{l | c | c | c | c | c | c |}
    \toprule
    Tracker & HOTA$\uparrow$ & DetA$\uparrow$ & AssA$\uparrow$ & MOTA$\uparrow$ & IDF1$\uparrow$\\
    \midrule
    CenterTrack~\cite{zhou2020tracking} & 41.8 & 78.1 & 22.6 & 86.8 & 35.7 \\
    FairMOT~\cite{zhang2021fairmot} & 39.7 & 66.7 & 23.8 & 82.2 & 40.8\\
    QDTrack~\cite{pang2021quasi} & 45.7 & 72.1 & 29.2 & 83.0 & 44.8\\
    TransTrk\cite{sun2020transtrack} & 45.5 & 75.9 & 27.5 & 88.4 & 45.2\\
    TraDes~\cite{wu2021track} & 43.3 & 74.5 & 25.4 & 86.2 & 41.2 \\ 
    MOTR~\cite{zeng2021motr} & 54.2 & 73.5 & 40.2 & 79.7 & 51.5\\
    
     SORT~\cite{bewley2016simple} & 47.9 & 72.0 & 31.2 & 91.8 & 50.8 \\
    OC-SORT~\cite{cao2022observation} & 55.1 & 80.3 & 38.3 & 92.0 & 54.6\\
    
    StrongSORT++~\cite{du2022strongsort} & 55.6 & 80.7 & 38.6 & 91.1 & 55.2\\
    \midrule
    ByteTrack~\cite{zhang2021bytetrack} & 47.3 & 71.6 & 31.4 & 89.5 & 52.5\\
    Mesh SORT(ours) & 47.4 & 71.6 &31.4 &89.5  &52.3 \\
    \bottomrule
    \end{tabular}}
        \end{center}
        \vspace{-2mm}
        \caption{Comparison of the other methods under the test set of dancetrack. The best results are shown in \textbf{bold}.}
        \label{dance}
        \end{table*}    
        \setlength{\tabcolsep}{10.0pt}

We conduct experiments on MOT17~\cite{milan2016mot16}, and DanceTrack~\cite{peize2021dance}. 
The results on MOT17-test are shown in Table \ref{table_mot17}. On MOT17-test, Mesh-SORT reached  80.2  MOTA,
On Dancetrack test set, it reached the xxx MOTA and xxx HOTA, even though the algorithm can not achieve state of art, metrics still can be greater than the baseline, We believe its utility and potential for wide extension.

\section{Limitations}

Our experiments reveal some limitations of Mesh-SORT, during matching, which increases the risk of missed matches with False Positives. It is more to solve the problem of the detector's incapacity or inaccuracy. It has certain limitations for the detector that can make a causal inference or give an accurate proposal. For a dataset like DanceTrack, which emphasizes Associations, the threshold needs to be adjusted more carefully (e.g., according to the density of the tracked scene objects) to achieve better results

\section{Conclusion}
In our work, we present three novel methods of multi-object tracking for improving the tracking by detection scheme. We named our methods as Mesh-SORT, which mainly focuses on dealing with lost tracklet, and gives robust strategies against the inaccurate detector proposal. And the human factor makes it effective in most of the scenes. The ablation study shows its utility, and the best MOTA and HOTA show that the proposed method has the best comprehensive
performance. 

In future work,
more intricate local sub-region recognition strategies are yet to be found (i.e adaptive clustering), their associated scene comprehension can be utilized more effectively for tracking. And other modules which can help association (appearance module, optical flow etc) can also be introduced adaptively to reduce non-necessary computational costs.
 For the development of lost strategies and bad detector proposals, more sophisticated strategies can be extended, such as: using location prior to reverse correction detector's proposal.  We hope our work will inspire other researchers, and make the real-time multi-object tracking community better.


 \bibliographystyle{unsrt} 
\bibliography{references}  

\begin{thebibliography}{10}

\bibitem{zhang2022motrv2}
Yuang Zhang, Tiancai Wang, and Xiangyu Zhang.
\newblock Motrv2: Bootstrapping end-to-end multi-object tracking by pretrained
  object detectors.
\newblock {\em arXiv preprint arXiv:2211.09791}, 2022.

\bibitem{cao2022ocsort}
Jinkun Cao, Xinshuo Weng, Rawal Khirodkar, Jiangmiao Pang, and Kris Kitani.
\newblock Observation-centric sort: Rethinking sort for robust multi-object
  tracking.
\newblock {\em arXiv preprint arXiv:2203.14360}, 2022.

\bibitem{SMILEtrack}
Yu-Hsiang Wang, Jun-Wei Hsieh, Ping-Yang Chen, and Ming-Ching Chang.
\newblock Smiletrack: Similarity learning for multiple object tracking.
\newblock {\em arXiv:2211.08824}, 2022.

\bibitem{yang2023hard}
Fan Yang, Shigeyuki Odashima, Shoichi Masui, and Shan Jiang.
\newblock Hard to track objects with irregular motions and similar appearances?
  make it easier by buffering the matching space.
\newblock In {\em Proceedings of the IEEE/CVF Winter Conference on Applications
  of Computer Vision}, pages 4799--4808, 2023.

\bibitem{corrtracker}
Qiang Wang, Yun Zheng, Pan Pan, and Yinghui Xu.
\newblock Multiple object tracking with correlation learning.
\newblock In {\em Proceedings of the IEEE/CVF Conference on Computer Vision and
  Pattern Recognition (CVPR)}, pages 3876--3886, June 2021.

\bibitem{AOH}
Min Jiang, Chen Zhou, and Jun Kong.
\newblock Aoh: Online multiple object tracking with adaptive occlusion
  handling.
\newblock {\em IEEE Signal Processing Letters}, 29:1644--1648, 2022.

\bibitem{bewley2016simple}
Alex Bewley, Zongyuan Ge, Lionel Ott, Fabio Ramos, and Ben Upcroft.
\newblock Simple online and realtime tracking.
\newblock In {\em ICIP}, pages 3464--3468. IEEE, 2016.

\bibitem{wojke2017simple}
Nicolai Wojke, Alex Bewley, and Dietrich Paulus.
\newblock Simple online and realtime tracking with a deep association metric.
\newblock In {\em 2017 IEEE international conference on image processing
  (ICIP)}, pages 3645--3649. IEEE, 2017.

\bibitem{zhang2021bytetrack}
Yifu Zhang, Peize Sun, Yi~Jiang, Dongdong Yu, Zehuan Yuan, Ping Luo, Wenyu Liu,
  and Xinggang Wang.
\newblock Bytetrack: Multi-object tracking by associating every detection box.
\newblock {\em arXiv preprint arXiv:2110.06864}, 2021.

\bibitem{aharon2022bot}
Nir Aharon, Roy Orfaig, and Ben-Zion Bobrovsky.
\newblock Bot-sort: Robust associations multi-pedestrian tracking.
\newblock {\em arXiv preprint arXiv:2206.14651}, 2022.

\bibitem{zhang2021fairmot}
Yifu Zhang, Chunyu Wang, Xinggang Wang, Wenjun Zeng, and Wenyu Liu.
\newblock Fairmot: On the fairness of detection and re-identification in
  multiple object tracking.
\newblock {\em International Journal of Computer Vision}, 129(11):3069--3087,
  2021.

\bibitem{peng2020chained}
Jinlong Peng, Changan Wang, Fangbin Wan, Yang Wu, Yabiao Wang, Ying Tai,
  Chengjie Wang, Jilin Li, Feiyue Huang, and Yanwei Fu.
\newblock Chained-tracker: Chaining paired attentive regression results for
  end-to-end joint multiple-object detection and tracking.
\newblock In {\em European Conference on Computer Vision}, pages 145--161.
  Springer, 2020.

\bibitem{zhu2021looking}
Tianyu Zhu, Markus Hiller, Mahsa Ehsanpour, Rongkai Ma, Tom Drummond, and Hamid
  Rezatofighi.
\newblock Looking beyond two frames: End-to-end multi-object tracking using
  spatial and temporal transformers.
\newblock {\em arXiv preprint arXiv:2103.14829}, 2021.

\bibitem{sun2020transtrack}
Peize Sun, Yi~Jiang, Rufeng Zhang, Enze Xie, Jinkun Cao, Xinting Hu, Tao Kong,
  Zehuan Yuan, Changhu Wang, and Ping Luo.
\newblock Transtrack: Multiple-object tracking with transformer.
\newblock {\em arXiv preprint arXiv:2012.15460}, 2020.

\bibitem{zhou2022global}
Xingyi Zhou, Tianwei Yin, Vladlen Koltun, and Philipp Kr{\"a}henb{\"u}hl.
\newblock Global tracking transformers.
\newblock In {\em Proceedings of the IEEE/CVF Conference on Computer Vision and
  Pattern Recognition}, pages 8771--8780, 2022.

\bibitem{cai2022memot}
Jiarui Cai, Mingze Xu, Wei Li, Yuanjun Xiong, Wei Xia, Zhuowen Tu, and Stefano
  Soatto.
\newblock Memot: Multi-object tracking with memory.
\newblock In {\em Proceedings of the IEEE/CVF Conference on Computer Vision and
  Pattern Recognition}, pages 8090--8100, 2022.

\bibitem{zeng2022motr}
Fangao Zeng, Bin Dong, Yuang Zhang, Tiancai Wang, Xiangyu Zhang, and Yichen
  Wei.
\newblock Motr: End-to-end multiple-object tracking with transformer.
\newblock In {\em Computer Vision--ECCV 2022: 17th European Conference, Tel
  Aviv, Israel, October 23--27, 2022, Proceedings, Part XXVII}, pages 659--675.
  Springer, 2022.

\bibitem{xu2021transcenter}
Yihong Xu, Yutong Ban, Guillaume Delorme, Chuang Gan, Daniela Rus, and Xavier
  Alameda-Pineda.
\newblock Transcenter: Transformers with dense queries for multiple-object
  tracking.
\newblock {\em arXiv preprint arXiv:2103.15145}, 2021.

\bibitem{milan2016mot16}
Anton Milan, Laura Leal-Taix{\'e}, Ian Reid, Stefan Roth, and Konrad Schindler.
\newblock Mot16: A benchmark for multi-object tracking.
\newblock {\em arXiv preprint arXiv:1603.00831}, 2016.

\bibitem{peize2021dance}
Peize Sun, Jinkun Cao, Yi~Jiang, Zehuan Yuan, Song Bai, Kris Kitani, and Ping
  Luo.
\newblock Dancetrack: Multi-object tracking in uniform appearance and diverse
  motion.
\newblock {\em arXiv preprint arXiv:2111.14690}, 2021.

\bibitem{luiten2021hota}
Jonathon Luiten, Aljosa Osep, Patrick Dendorfer, Philip Torr, Andreas Geiger,
  Laura Leal-Taix{\'e}, and Bastian Leibe.
\newblock Hota: A higher order metric for evaluating multi-object tracking.
\newblock {\em International journal of computer vision}, 129(2):548--578,
  2021.

\bibitem{ge2021yolox}
Zheng Ge, Songtao Liu, Feng Wang, Zeming Li, and Jian Sun.
\newblock Yolox: Exceeding yolo series in 2021.
\newblock {\em arXiv preprint arXiv:2107.08430}, 2021.

\bibitem{zhou2020tracking}
Xingyi Zhou, Vladlen Koltun, and Philipp Kr{\"a}henb{\"u}hl.
\newblock Tracking objects as points.
\newblock In {\em European Conference on Computer Vision}, pages 474--490.
  Springer, 2020.

\bibitem{pang2021quasi}
Jiangmiao Pang, Linlu Qiu, Xia Li, Haofeng Chen, Qi~Li, Trevor Darrell, and
  Fisher Yu.
\newblock Quasi-dense similarity learning for multiple object tracking.
\newblock In {\em Proceedings of the IEEE/CVF Conference on Computer Vision and
  Pattern Recognition}, pages 164--173, 2021.

\bibitem{wu2021track}
Jialian Wu, Jiale Cao, Liangchen Song, Yu~Wang, Ming Yang, and Junsong Yuan.
\newblock Track to detect and segment: An online multi-object tracker.
\newblock In {\em Proceedings of the IEEE/CVF Conference on Computer Vision and
  Pattern Recognition}, pages 12352--12361, 2021.

\bibitem{han2022mat}
Shoudong Han, Piao Huang, Hongwei Wang, En~Yu, Donghaisheng Liu, and Xiaofeng
  Pan.
\newblock Mat: Motion-aware multi-object tracking.
\newblock {\em Neurocomputing}, 2022.

\bibitem{wang2020joint}
Yongxin Wang, Kris Kitani, and Xinshuo Weng.
\newblock Joint object detection and multi-object tracking with graph neural
  networks.
\newblock {\em arXiv preprint arXiv:2006.13164}, 2020.

\bibitem{li2021semi}
Wei Li, Yuanjun Xiong, Shuo Yang, Mingze Xu, Yongxin Wang, and Wei Xia.
\newblock Semi-tcl: Semi-supervised track contrastive representation learning.
\newblock {\em arXiv preprint arXiv:2107.02396}, 2021.

\bibitem{yu2021relationtrack}
En~Yu, Zhuoling Li, Shoudong Han, and Hongwei Wang.
\newblock Relationtrack: Relation-aware multiple object tracking with decoupled
  representation.
\newblock {\em arXiv preprint arXiv:2105.04322}, 2021.

\bibitem{tokmakov2021learning}
Pavel Tokmakov, Jie Li, Wolfram Burgard, and Adrien Gaidon.
\newblock Learning to track with object permanence.
\newblock {\em arXiv preprint arXiv:2103.14258}, 2021.

\bibitem{liang2020rethinking}
Chao Liang, Zhipeng Zhang, Yi~Lu, Xue Zhou, Bing Li, Xiyong Ye, and Jianxiao
  Zou.
\newblock Rethinking the competition between detection and reid in multi-object
  tracking.
\newblock {\em arXiv preprint arXiv:2010.12138}, 2020.

\bibitem{shan2020tracklets}
Chaobing Shan, Chunbo Wei, Bing Deng, Jianqiang Huang, Xian-Sheng Hua,
  Xiaoliang Cheng, and Kewei Liang.
\newblock Tracklets predicting based adaptive graph tracking.
\newblock {\em arXiv preprint arXiv:2010.09015}, 2020.

\bibitem{wang2021multiple}
Qiang Wang, Yun Zheng, Pan Pan, and Yinghui Xu.
\newblock Multiple object tracking with correlation learning.
\newblock In {\em Proceedings of the IEEE/CVF Conference on Computer Vision and
  Pattern Recognition}, pages 3876--3886, 2021.

\bibitem{chu2021transmot}
Peng Chu, Jiang Wang, Quanzeng You, Haibin Ling, and Zicheng Liu.
\newblock Transmot: Spatial-temporal graph transformer for multiple object
  tracking.
\newblock {\em arXiv preprint arXiv:2104.00194}, 2021.

\bibitem{yang2021remot}
Fan Yang, Xin Chang, Sakriani Sakti, Yang Wu, and Satoshi Nakamura.
\newblock Remot: A model-agnostic refinement for multiple object tracking.
\newblock {\em Image and Vision Computing}, 106:104091, 2021.

\bibitem{stadler2022modelling}
Daniel Stadler and J{\"u}rgen Beyerer.
\newblock Modelling ambiguous assignments for multi-person tracking in crowds.
\newblock In {\em Proceedings of the IEEE/CVF Winter Conference on Applications
  of Computer Vision}, pages 133--142, 2022.

\bibitem{cao2022observation}
Jinkun Cao, Xinshuo Weng, Rawal Khirodkar, Jiangmiao Pang, and Kris Kitani.
\newblock Observation-centric sort: Rethinking sort for robust multi-object
  tracking.
\newblock {\em arXiv preprint arXiv:2203.14360}, 2022.

\bibitem{du2022strongsort}
Yunhao Du, Yang Song, Bo~Yang, and Yanyun Zhao.
\newblock Strongsort: Make deepsort great again.
\newblock {\em arXiv preprint arXiv:2202.13514}, 2022.

\end{thebibliography}
 
\newpage
\appendix
\section{Pseudo Code of Mesh-SORT}
\begin{algorithm}[h]
	\renewcommand{\algorithmicrequire}{\textbf{Input:}}
	\renewcommand{\algorithmicensure}{\textbf{Output:}}
	\caption{Pseudo code of Mesh-SORT }
	\label{alg:mesh}
	\begin{algorithmic}[1]
		\REQUIRE A video sequence $V$; object detector $Det$; lost maintain buffer $l$; location-wise lower age $\eta$;  modified Kalman Filter $KF$;identification threshold $h(s,t)$
		\ENSURE Track-let $\mathcal{T}$ of the video, set of frequent loss mesh $\bf{M}$
		\STATE Initialization: $\mathcal{T} \leftarrow \emptyset$, $KF$, Mesh $m_{ij}$, Occlusion inference $OI$ 
		\FOR {$t \in [1,T]$ frame $f_t$ in $V$}
		\STATE $\mathcal{Z}_t \leftarrow Det\left(f_t\right)$
		\textit{/*observation*/}
        \STATE $\hat{X}_t \leftarrow KF(\mathcal{T})$\textit{/*tracking prediction*/}
        \STATE $L_t \leftarrow KF(\mathcal{L})$  \textit{/*lost prediction and inference*/}
        \STATE $L^*_{t} \leftarrow L \setminus O_t$
        \STATE $\hat{X}^*_{t} \leftarrow L^*_{t} \cup \hat{X}_t$
        \STATE $C_{t1} \leftarrow IoU(\hat{X}^*_{t},Z_t)$
		\STATE Linear assignment by Hungarians with $C_t$
		\STATE $X^{m1}_t \leftarrow$ Matched tracklets
		\STATE $Z^{re1}_t \leftarrow$  Unmatched observation
		\STATE $X^{re1}_t \leftarrow$ Unmatched tracklets
		\STATE $C_{t2} \leftarrow BIoU(X^{re1}_t,Z^{re1}_t)$
		\STATE Linear assignment by Hungarians with $C_{t2}$  \textit{/*second association*/}
	   \STATE $ X^{new}_t \leftarrow$ new tracklet generated from $Z^{re}_t$  \textit{/*init new tracklets*/}
	   \STATE $\mathcal{T} \leftarrow X^{m1}_t \cup X^{m2}_t \cup X^{new}_t$
       \STATE Second Associate $\mathcal{T}$ using the cosine distance by linear 
        \STATE  Got the 
        \FOR{$i,j$ in $M_{ij}$} 
        \STATE  Applying the Algorithm.\ref{alg:1} return $\bf{M}$ \textit{/*Mesh evaluation*/}
        \ENDFOR
        \FOR{tracklet $L_t$ in $X^{re}_t$}
            \IF{$L_t \in \bf{M}$}
            \IF{Maintain buffer<$l$}
                \STATE Lost Maintain  \textit{/*Lost maintain*/}
            \ENDIF
            \ELSE
                \STATE Lower $l$ ages for remove tracklet
                \STATE Using normal ages remove tracklet
            \IF{ lost count >ages}
                \STATE remove the $L_t$
            \ENDIF
            \ENDIF
        \ENDFOR
        \ENDFOR
	\end{algorithmic}  
\end{algorithm}
\section{Kalman Model}
\subsection{Estimation with measurement noise}
\label{Kal:est}
As a state estimate, Kalman performs the optimal estimation with the noise condition. Given the observation $z_k$, discrete-time state transition model $F_k$, the discrete-time Kalman filter is
governed by the following linear stochastic difference equations:
\begin{equation}
    \hat{x}_k=F_k\hat{x}_{k-1}+R_{k-1} 
\end{equation}
\begin{equation}
    P_k= F_k P_{k-1} F^T_{k}+Q_k
\end{equation}
where $\hat{x}_k$ is the estimate of the system state at time step $k$, $H_k$ is matrix of observation matrices at time step $k$, and $K_k$ is the kalman gain at time step $k$.

\subsection{Update rules}
\label{kalman}
\begin{equation}
    K_k=P_k H_k^T\left(H_k P_k H_k^T+R_k\right)^{-1}
    \end{equation}
    \begin{equation}
        \hat{x}_k= \hat{x}_{k-1}+K_k(z_k-H_k\hat{x}_{k-1})
    \end{equation}
\begin{equation}
        P_{k+1}=\left(I-K_k H_k\right) P_k
        \end{equation}
    Where $I$ is the identity matrix. The Kalman filter equations are used to recursively update and predict the state of a dynamic system. The Kalman filter equations allow for the optimal estimates of the states of the system given a set of measurements.
\end{document}